\pdfoutput=1

\documentclass[11pt]{article}

\usepackage{ACL2023}
\usepackage{times}
\usepackage{latexsym}
\usepackage{tabularx}
\usepackage{booktabs}
\usepackage{graphics}
\usepackage{xcolor}

\usepackage{svg}
\usepackage{rotating}
\usepackage[T1]{fontenc}
\usepackage[utf8]{inputenc}

\usepackage{microtype}

\usepackage{inconsolata}
\usepackage{amssymb}%
\usepackage{pifont}%

\title{Byte-Level Grammatical Error Correction\\ Using Synthetic and Curated Corpora}

\newcommand{\mideind}{{ \mathrm{1}}}
\newcommand{\pioneer}{{ \mathrm{2}}}

  \author{Svanhvít Lilja Ingólfsdóttir$^{\mideind}$~\;~Pétur Orri Ragnarsson$^1$~\;~Haukur Páll Jónsson$^{\mideind}$ \\
  \bf{Haukur Barri Símonarson$^{\mideind}$~\;~Vilhjálmur Þorsteinsson$^{\mideind}$~\;~ Vésteinn Snæbjarnarson$^{{\mideind},{\pioneer}}$ }  \\
  $^{\mideind}$Miðeind ehf.
  ~\;~ $^{\pioneer}$University of Copenhagen\\
  \texttt{\{\href{mailto:svanhvit@mideind.is}{svanhvit}, \href{mailto:petur@mideind.is}{petur}, \href{mailto:haukurpj@mideind.is}{haukurpj}, \href{mailto:haukur@mideind.is}{haukur}, \href{mailto:vt@mideind.is}{vt}, \href{mailto:vesteinn@mideind.is}{vesteinn}}\}@mideind.is
  }
\begin{document}
\maketitle
\begin{abstract}
Grammatical error correction (GEC) is the task of correcting typos, spelling, punctuation and grammatical issues in text. Approaching the problem as a sequence-to-sequence task, we compare the use of a common subword unit vocabulary and byte-level encoding. Initial synthetic training data is created using an error-generating pipeline, and used for finetuning two subword-level models and one byte-level model. Models are then finetuned further on hand-corrected error corpora, including texts written by children, university students, dyslexic and second-language writers, and evaluated over different error types and origins. We show that a byte-level model enables higher correction quality than a subword approach, not only for simple spelling errors, but also for more complex semantic, stylistic and grammatical issues. In particular, initial training on synthetic corpora followed by finetuning on a relatively small parallel corpus of real-world errors helps the byte-level model correct a wide range of commonly occurring errors. Our experiments are run for the Icelandic language but should hold for other similar languages, particularly morphologically rich ones.

\vspace{0.5em}
\hspace{.5em}\includegraphics[width=1.25em,height=1.25em]{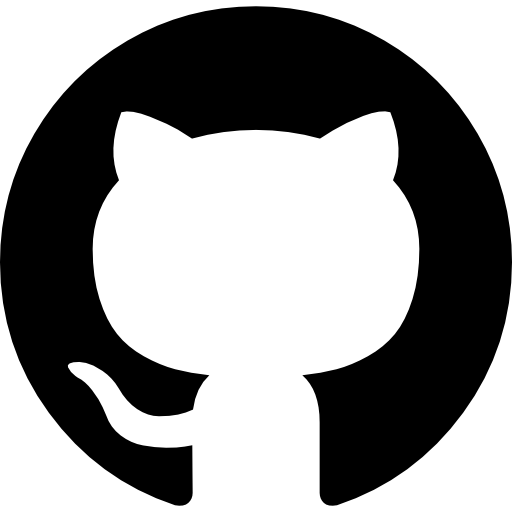}\hspace{.75em}\parbox{\dimexpr\linewidth-5\fboxsep-2\fboxrule}{\url{https://github.com/mideind/byte-gec}}
\vspace{-.5em}
\end{abstract}

\begin{figure*}[h]
    \centering
    \includesvg[width=\textwidth]{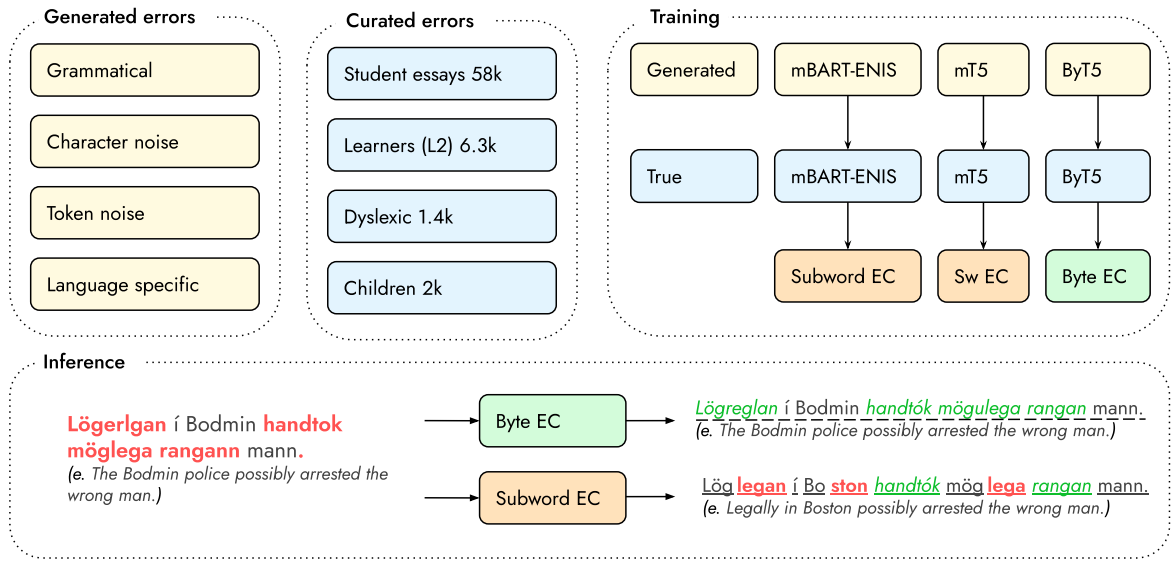}
    \caption{Overview of training data and comparison of output. The Icelandic-English mBART-ENIS model and the multilingual ByT5 and T5 models are first trained on generated parallel error corpora before being adapted on curated (collected) true error corpora in Icelandic. The final models are compared on an error correction (EC) task in Icelandic. The example demonstrates how the byte-level model performs well while the subword model cannot see the individual characters in every word, leading to degraded performance. %
    }
    \label{fig:hero}
\end{figure*}

\section{Introduction}

Spelling mistakes due to typos and rushed writing, nonstandard punctuation and spelling, and grammatical and stylistic issues are common to almost everyone who writes any kind of text. This applies in any language and can distract the reader or make the communication miss its mark. 
This can hinder people who have difficulties writing text conforming to a particular language standard, be it due to disability, dyslexia, linguistic background, limited access to education or any other reason. Prejudice against people whose writing deviates from the standard can make some shy away from communicating with others, leaving their voices out of important discussions and restricting their opportunities \cite{alexander-passe-2015-dyslexia}.

Grammatical error correction (GEC) is the task of adjusting a text's spelling, grammar, and linguistic style to conform to an approved language standard or convention \cite{rauf-2017-automated}. %
While the latest work on GEC is based on Transformer models \cite{attention}, the subword tokenization methods commonly used in these models are a source of problems when it comes to typos and other variants \cite{schmaltz-etal-2017-adapting}. Subword tokenization \cite{sennrich-etal-2016-neural,kudo-2018-subword} was presented as a solution to the open vocabulary problem, as it is a compromise between character encoding and whole word encoding, enabling unknown words to be represented using known subwords. 

However, a significant downside of subword tokenization is how it is affected by noisy input; if a word contains a typo or other spelling variants, this can completely shift its representation. In addition, in languages with rich morphology, a word can have many different surface forms, some rarer than others, that all carry the meaning of the base word, but appear in different syntactic contexts. A subword-tokenized model may struggle to capture the nuances of such a language effectively since it may need several different subwords to represent a single word, depending on spelling and context. When an unfamiliar variant of the word appears in unseen text, the model is challenged to decode it correctly, even when it results in uncommon subword units. 

Our motivation is that a byte or character-level approach should intuitively be more robust to spelling or morphology variations, as it is not constrained by the subword vocabulary. We explore using a byte-level architecture, ByT5 \cite{xue-2022-byt5}, for correcting everything from typos to complex grammatical issues in text. The language studied is Icelandic, a highly-inflected North Germanic language. For instance, the morphological complexity in Icelandic means that nouns can have up to 16 different surface forms, and adjectives over 50. GEC for a morphologically complex language needs to go beyond correcting only single words or limited phrases; it needs to consider the syntax of the whole sentence. This is the case for Icelandic but also for other languages with rich morphology, such as Arabic, Hebrew, Polish, Basque, Lithuanian and Hungarian, to name a few.

We compare the performance of the byte-level architecture to two subword-based architectures; ByT5's predecessor, mT5 \cite{xue-2021-mt5}, and an mBART \cite{liu-etal-2020-multilingual-denoising} model that has been pretrained further on both Icelandic and English. We employ real and synthetic error data for training, and present models and a framework for error generation methods that can be adapted to other languages. For under-resourced languages such as Icelandic \cite{rognvaldsson-2022-report}, using synthetic training data makes neural training for GEC a viable option. %

Our main contributions include a comparison between subword tokenization and byte-level tokenization for GEC when training over a combination of curated and synthesized data. We demonstrate how byte-level models not only bypass subword-related issues, but can also correct long-range errors in text. We release our error generation framework as well as models for GEC using byte-level and subword tokens in Icelandic.
While our work focuses on the Icelandic language, we have no reason to believe that similar results do not hold for other languages, particularly those similar to Icelandic in terms of morphological complexity.

\section{Related work}
The bulk of research on grammatical error detection and correction has been focused on English and English learner texts, due to existing training data and benchmarks, and the large market of English learners worldwide who benefit from an automatic language correction tool \cite{naplava-straka-2019-grammatical}. While spelling and grammar errors appear in every language, each language has its own set of error types that are more common than others, due to different phonetic, morphological and syntactic characteristics.

\subsection{Synthetic data generation for GEC}
The problem of data scarcity in GEC, when approached as a sequence-to-sequence task, is typically addressed with synthetic data generation \cite{stahlberg-kumar-2021-synthetic}.
One approach to creating ungrammatical sentences uses random character noise and simple rules to manipulate the text. Another approach is using a spell checker in reverse to noise text \cite{grundkiewicz-junczys-dowmunt-2019-minimally}. In contrast, others have used probabilities derived from an annotated corpus of naturally occurring errors to corrupt text \cite{felice-2014-generating}. Recent efforts widely employ neural networks to create synthetic errors \cite{stahlberg-kumar-2021-synthetic}; many use methods derived from machine translation \cite{junczys-etal-2018-approaching}, for example, by creating worse text using deliberately bad translation models \cite{xie-etal-2018-noising, zhou-etal-2020-improving-grammatical} or roundtrip translations between languages \cite{lichtarge-etal-2019-corpora}. Yet another option is to leverage available resources with edits, such as Wikipedia edit histories, to generate corrupted corpora \cite{grundkiewicz-2014-wiked}. %

\subsection{Sequence segmentation for GEC}
\label{sec:problem}
GEC can essentially be considered the task of generating grammatical target text from an ungrammatical source, similar to machine translation. The idea of approaching GEC as a machine translation problem dates back to 2006 \cite{brockett-etal-2006-correcting}, and this approach has since become the most prevalent method of GEC, with the focus shifting from statistical machine translation (SMT) to neural methods as they developed \cite{yuan-briscoe-2016-grammatical, ji-etal-2017-nested, schmaltz-etal-2017-adapting, chollampatt-2018-multilayer, junczys-etal-2018-approaching}. However, phrase-based SMT continued to be the state-of-the-art for GEC for longer than in the field of interlingual neural machine translation (NMT) \cite{junczys-etal-2018-approaching}. This is partly because of the data scarcity problem and partly because of the tokenization methods typically used in transformer models. Breaking words up into subword units decreases the vocabulary size while addressing the out-of-vocabulary problem \cite{sennrich-etal-2016-neural}. A prominent drawback of this approach is that the fixed subword vocabulary makes the models sensitive to noise in the text \cite{tay-2021-charformer, eger-benz-2020-hero}. 

In a subword-based GEC model, when a word contains a typo or is spelled unconventionally, it may look like an unknown word, for which no known representation exists. The model may then segment the word differently from what was seen during training, causing mispredictions. If the subword representation for ``different'' is \texttt{[\_diff, er, ent]}, but the word is misspelled as ``diffirent'', its subwords might be \texttt{[\_diffi, ren, t]}, and a subword-based GEC model might correct the typo by outputting a different word, \texttt{[\_diffi, cul, t]}. This issue is highlighted in Figure \ref{fig:hero}, also showing how a byte-based approach is not limited by this issue. %

This is also true for unseen words that are correctly spelled, such as foreign-named entities, which can lead to the subword-based GEC model ``correcting'' a perfectly spelled word it has not seen before, by replacing it with the most likely candidate. In the sentence ``The tournament was held in Espoo, Finland.'', the place name ``Espoo'' may be represented by a single subword token \textit{Espoo}. Since this token is unfamiliar, the model finds the most likely subword token for this particular sentence, \textit{Helsinki}.\footnote{Actual example from our evaluation data, generated by an mBART subword model.} This changes the semantics of the sentence and can introduce serious errors. The result is a grammatically correct and meaningful sentence, but the semantics have drifted away from the original text.

Due to this known shortcoming of subword tokenization \cite{schmaltz-etal-2017-adapting}, efforts have been made to design architectures where the characters or the underlying bytes are used directly as input tokens. Byte and character-level models inevitably result in much longer sequences than subword models, making them more costly to train, and slower in inference. Some truly token-free general-purpose architectures that are increasingly competitive to token-based models have emerged recently, including CANINE \cite{clark-2022-canine} (character-level), PIXEL \cite{pixel} (text-to-image),
and ByT5 \cite{xue-2022-byt5} (byte-level). The training of ByT5 is based on the subword-based multilingual mT5 \cite{xue-2021-mt5} approach, but in comparison, the model is equipped with a heavier encoder (three times the depth of the decoder). Compared to mT5, ByT5 is more robust to noisy input, but inference is slower (1.5 to 2.6 times slower on average on a transliteration task, and up to 9 times longer on tasks with longer input sequences) \cite{xue-2022-byt5}. Another approach to making subword models more robust is using subword regularization to produce multiple segmentations of the same word \cite{kudo-2018-subword, provilkov-etal-2020-bpe}. This is commonly used for addressing the open vocabulary problem and noisy data, such as ungrammatical text.

Despite the reported advantage of character or byte-based Transformer models on noisy text \cite{libovicky-etal-2022-dont}, work using this approach to GEC is not very common.
A notable exception is work for GEC in the Lithuanian language, where ByT5 has been used for diacritics restoration \cite{stankevicius-2022-correcting} and limited GEC \cite{stankevicius-2022-towards}. They generate synthetic data by applying noise (common typographical errors, swapping letters for similar sounding ones, other non-grammatical word-level noise) to a crawled and filtered corpus and compare results when training with T5 and ByT5. Their findings agree with ours that the byte-level model outperformed the subword model.

Our work deviates from that of \citet{stankevicius-2022-correcting} in the following key ways: We (a) generate more sophisticated and realistic errors using grammatical information from part-of-speech (PoS) tags and using custom rules based on empirical findings; (b) combine generated error data with true error corpora from a wide range of demographic sources; and (c) explore in detail which error and text types benefit the most from finetuning on true error corpora, as opposed to training on synthetic data. As far as we are aware, no attempt at byte-level Transformer-based GEC, trained on synthetic and real error corpora, has been published.
Concurrent work using ByT5 for Icelandic is \cite{jasonarson-etal-2023-generating}, where errors in the OCR output of historical Icelandic texts are corrected using a generated corpus of errors extracted from real data.

Current state-of-the-art in GEC is based on sequence-tagging methods \cite{omelianchuk-etal-2020-gector}, which instead of generating whole sequences, tag the erroneous sentence with its corrections, and on sequence-to-sequence methods, as has been described. Further work has explored automatic character transformations for GEC tagging \cite{straka-etal-2021-character} to better handle character-level errors. One of the current highest-scoring models on English GEC benchmarks is gT5 \cite{rothe-etal-2021-simple}, which is based on the mT5 model. %

\subsection{Prior work for Icelandic}
Apart from some rule-based spell checkers that don't make use of the full context, one rule-based correction system exists for Icelandic, based around parse trees, GreynirCorrect \cite{oladottir-etal-2022-developing}. This system is contingent on the sentence parsing according to a pre-defined context-free grammar, and can only handle issues that fit pre-defined rules. This setup is both a strength and a weakness as the system is highly configurable and capable of many things, such as detecting syntactic inconsistencies and errors, and can give the user useful information on the errors found. Still, when a text has many errors, complexity builds up and rules can start interfering. Sentences containing many issues, such as from users with dyslexia, generally have lower accuracy using this method. 

The work presented here is the first where neural networks are used in GEC for Icelandic. \citet{snaebjarnarson-etal-2022-warm} use neural methods for detecting such errors, but not for correcting them.

\section{Methods}

\subsection{Curated dataset}
\label{sec:curated}
A single collection of parallel error corpora exists for Icelandic, the Icelandic Error Corpus (IceEC).
The corpora are annotated and corrected by language experts \cite{arnardottir-2021-creating}. The dataset is highly granular in its categories, containing hundreds of labels, but with a limited number of high-level groups (\emph{coherence, grammar, orthography, style} and \emph{vocabulary}).

The IceEC is split into a larger general corpus and three specialized corpora \cite{arnardottir-etal-2022-error}. The general one contains 58,239 sentences of student essays, online news texts and Wikipedia articles. This corpus is annotated with around 50k errors of different categories. The three specialized corpora are much smaller and contain texts from Icelandic language learners (6270 sentences), dyslexic native speakers (1362 sentences), and children (2070 sentences), volunteered by the users themselves. These smaller corpora contain more errors per sentence than the general one, and add diversity to the training data.

This curated error data was used for finetuning our models, and combined into one training dataset for a total of 64k input sequences (single sentences), after setting aside validation and test data. The general IceEC also includes a 5.3k sentence test set used for evaluation.

\subsection{Synthetic dataset}

\begin{table}[htbp!]
\small
\centering
\begin{tabularx}{\columnwidth}{lll}
\toprule
\textbf{Noise} \\
\midrule
1. Grammatical case swapped in nominals* \\
2. Indicative changed to subjunctive* \\
3. Dativitis* \\ 
4. Spaces added to words according to morphological\\ \hspace{12pt}rules* \\
5. Known misspellings added  \\ 
6. Spaces deleted between words \\
7. Commas deleted from sentence \\
8. Word order swapped \\
9. Words duplicated \\
10. Characters duplicated \\
11. Characters dropped \\
12. Characters swapped according to simple \\ \hspace{12pt} language-specific rules (y $\leftrightarrow$ i, ýi$\rightarrow$ýji ...) \\
13. Characters accented or accents removed\\ \hspace{12pt} (a$\leftrightarrow$á, I$\leftrightarrow$Í ...) \\
14. Random character replacement \\
\bottomrule
\end{tabularx}
\normalfont
\caption{Noise and modifications used on a high-quality corpus (IGC) to generate synthetic training data. We use * to indicate noise that relies on PoS tags and/or morphological lookup.}
\label{tab:noise}
\end{table}

We applied a diverse set of methods for error generation, both using linguistic knowledge and random noising methods. This rule-based approach to synthetic data generation gave us control over the types of noise applied, and allowed us to generate evaluation data for each error type.

As our basis of correct text to be noised, we used the Icelandic Gigaword Corpus (IGC) \cite{steingrimsson-2018-very}, a collection of Icelandic editorial texts. These are mostly news articles, published literature and legal texts. We selected from this corpus those text sources that are the most likely to have been reviewed as part of the editorial process of each publication/source (literature, journals, news, laws, adjudications and transcribed parliamentary speeches). Some of these texts still have their share of typos and other errors and inconsistencies, especially the news articles, which were deemed important training data because of their general vocabulary, not found in the more formal text sources. As a preprocessing step, we filtered out lower-quality and irrelevant sentences, by removing sentences containing mostly foreign texts, illegal characters and words with known misspellings, sourced from lists of common misspellings. The corpus was tokenized using the Greynir Tokenizer \cite{thorsteinsson-etal-2019-wide} and PoS tagged using the GreynirSeq tagger \cite{snaebjarnarson-etal-2022-warm}.

We generated three categories of errors: 1) noise within words; 2) noise at the sequence level; and 3) grammatical and morphological modifications. The first two resemble those used when noising back-translation data \cite{edunov-2018-understanding}. The third type is based on using available tools and linguistic knowledge to create errors that are unlikely to be formed randomly, but resemble those of human writers. %

In order to explore to what extent subword and byte-level models can learn and generalize grammatically complex issues in a morphologically rich language, we go beyond naive language-agnostic noising of text. A more detailed explanation of the Icelandic-specific noise is given in Appendix \ref{sec:appendixis}. The noise methods are shown in Table \ref{tab:noise}. This is by no means a finite list of linguistic variants or errors found in Icelandic texts, but constitutes examples chosen for studying the model performance on these more grammatically complex challenges.

The error generator allows for noising levels to be configured via hyperparameters. Experiments with different noise ratios in the synthetic data showed that highly noised text provided the best training examples, without the models learning to ``overcorrect``, i.e., to introduce false positives. Instead of producing even more synthetic data, we geared up the noise to produce highly error-dense examples, setting the random and more naive error noise to appear in 80\% of cases, and the rule-based error noise to be used wherever possible.\footnote{This may seem excessive, but preliminary experiments with the mBART model showed that lower noise ratios resulted in too much copying behavior.} Examples of parallel sentences with and without synthetic errors are shown in Table \ref{tab:syntherrorexamples}. A total of 35M synthetic parallel sentences were generated using these methods. 2000/4000 lines were set aside for validation/testing, respectively, and special evaluation sets were generated for each error type (see Section \ref{sec:results}).

\begin{table}
\small
\begin{tabularx}{\columnwidth}{l X}
\toprule
\textbf{Source} & \textbf{Text}\\
\midrule
Synth & \textit{\textcolor{red}{Atvinnuleisi}} er ekki \textit{\textcolor{red}{eein afff a4f ástæðna fólksfækkun áNorðurlandi véstrá}}. \\ 
Orig & Atvinnuleysi er ekki ein af ástæðum fólksfækkunar á Norðurlandi vestra. \\
\midrule
Synth & Ég er \textit{\textcolor{red}{mað 4eita}} að \textit{\textcolor{red}{ogg vynnu mér langarað}} fá að skoða \textit{\textcolor{red}{atvinnuauglýsingrnar.."}} \\
Orig & Ég er að leita að vinnu og mig langar að fá að skoða atvinnuauglýsingarnar.
\\

\bottomrule
\end{tabularx}
\caption{Synthetic error examples.}
\label{tab:syntherrorexamples}
\end{table}

\subsection{Models}
We compared three model architectures to evaluate the differences between using subword tokenization and a byte-level method. %
Comparing models with different architectures calls for defining which factors are compared. In particular, byte sequences are longer than subword sequences when counting the number of tokens, roughly 4 times longer on average in the original multilingual mT5/ByT5 training data \cite{xue-2021-mt5}. And as \citet{xue-2022-byt5} note, the mT5 architecture tends to need less training time than ByT5. We compared the models after an equal amount of training samples (100k updates). We also continued the training of the ByT5 model, using more than five times the number of updates.

\subsubsection{mBART}
We continued training of the pretrained multilingual BART25 model (mBART) \cite{liu-etal-2020-multilingual-denoising}, using the original pretraining objective on texts in Icelandic and English. The training of the model is detailed in Appendix \ref{sec:mbart}. This model, mBART-ENIS, was then finetuned on the synthetic error data, to teach it to correct errors in Icelandic texts. %

Training on the synthetic error data was performed with an effective batch size of 3000 input tokens (roughly 60 sequences), a learning rate of \textrm{3e-5} with an inverse square root scheduler, 0.1 dropout, 0.3 attention dropout, 1000 warmup steps, 0.1 label smoothing, no weight decay, and using the Adam optimizer, for 100k updates on an A100 card for a day.\footnote{The model had not completely converged, but little gains were observed on the validation data when compared to training for 150k updates.} 

In addition to the above experiments, we conducted separate experiments using segmentation regularization \cite{kudo-2018-subword} to introduce more noise to the training examples and explore alternative measures to mitigate the subword tokenization problem. The BART architecture uses unigram subword units; we applied subword regularization with $\alpha = 0.7$ and keep all other parameters unchanged.

\subsubsection{ByT5}
For the byte-level approach we employed the ByT5-base model \cite{xue-2022-byt5}, which is based on the multilingual T5 model \cite{raffel-2020-exploring}, but operates on bytes instead of subwords. The ByT5 model is pretrained on over 100 languages, but has only seen a limited amount of Icelandic. The mC4 dataset which is used to train ByT5 and mT5 is also lacking in quality for low-resource languages, in particular for Icelandic, as shown by \citet{snaebjarnarson-etal-2022-warm}.

The pretraining task in ByT5 has been adapted to a byte-level model, with span infilling based on bytes, not subwords. Apart from this, the main difference between the mT5 and ByT5 model architecture is the heavier encoder of ByT5.%

Sequences in byte-level architectures are long and correspond more or less to the number of characters in Icelandic, resulting in increased training time. We trained the ByT5-base model using a maximum sequence length of 512 bytes, which was found to be a reasonable compromise, as most sentences in Icelandic texts are shorter than this.

The ByT5-base model was finetuned on the synthetic data with an effective batch size of 32 sequences (sentences). The learning rate was set to \textrm{2e-5} using the Adam optimizer with 1000 warmup steps and no weight decay. This model was further trained for a total of 550k updates, or 13 A100 card days.\footnote{Training was stopped after two weeks due to computing limitations, even though the validation loss was still decreasing. However, this potential undertraining is not crucial to our results, as we are mostly interested in evaluating the model at earlier stages. This should be taken into consideration in future work.}

\subsubsection{mT5}
For a more direct comparison of byte-level and subword-level models, we also finetuned the mT5-base \cite{xue-2021-mt5} model on the same data, with the notable difference to the mBART model that it was not further trained on Icelandic. The mT5 models were pretrained on the same data as ByT5 and have a similar architecture, as described above, and are thus as comparable as subword and byte-level models can be. As previously mentioned, mT5-base is the base for the state-of-the-art gT5 \cite{rothe-etal-2021-simple} model for multilingual GEC.

We finetuned the mT5-base model on the synthetic data using the same parameters as in our ByT5-base finetuning and evaluated it at 100k updates.\footnote{The models had not completely converged at this point, but we only saw marginal gains by adding 80k more steps to the synthetic finetuning.}

\subsection{Finetuning on curated corpora}
Using the curated error corpora (IceEC), we finetuned the byte-level and subword-level models to convergence. For the mBART model, this meant training with a learning rate of \textrm{2e-6} for 53k updates (67 epochs), with attention dropout set to 0.1, weight decay to 0.001 and other parameters being the same as during the synthetic finetuning.

The ByT5 and mT5 models were finetuned with a learning rate of \textrm{2e-6}, other parameters were the same as during finetuning on the synthetic data. The ByT5 model had converged at 120k updates (60 epochs), while the mT5 was still improving on the validation data at 200k updates (100 epochs), but with time we found it forgot too much of the synthetic error correction task. We report evaluation scores at 130k.

For comparison, we also finetuned the different models (mBART-ENIS, mT5 and ByT5-base) on the IceEC data only, without the synthetic finetuning phase. This was done to examine how much the models learn from the added synthetic examples, and how far we can get using a small amount of hand-corrected examples. The mT5 and ByT5 models were trained for 100k updates and the mBART-ENIS model for 10k updates.

\section{Results}
\label{sec:results}
Different metrics exist for evaluating GEC performance, but most are language-specific, and have not been adapted to Icelandic. Here we employ a language agnostic metric for scoring our models, the GLEU score \cite{napoles-etal-2015-ground, napoles-2016-gleu}. GLEU\footnote{Not to be confused with Google BLEU, also called GLEU \cite{wu-2016-googles}.} is a variant of the BLEU \cite{papineni-etal-2002-bleu} score used to evaluate machine-translation output. It has been modified to account for both the source and the reference, by rewarding overlap between the source and the target sentence, and penalizing n-grams that should have been changed, but were not.

When evaluating GEC for English, ERRANT \cite{bryant-etal-2017-automatic} is commonly used. It is a span-based correction method that uses the $F\textsubscript{0.5}$ metric, where precision weighs twice as much as recall. Though this metric has not been implemented for Icelandic, we also report ERRANT scores using a language-agnostic approach, disregarding error types and only reporting the span-based $F\textsubscript{0.5}$ scores for each test set. These results are shown in Table \ref{tab:errant-results} in Appendix \ref{sec:errant}; they align well with the GLEU results in Table \ref{tab:results} which are described below.

\begin{table*}
\small
\resizebox{\textwidth}{!}{%
\begin{tabularx}{1.15\textwidth}{l|l||lllll|llllllll}
\toprule
 \multicolumn{1}{c}{} & & \multicolumn{5}{c|}{\sc{Curated Error Corpora}} & \multicolumn{8}{c}{\sc{Synthetic Error Corpora}} \\ 
\midrule 
\sc{Training Data} & \sc{Model} & \rotatebox{90}{IceEC.test} &  \rotatebox{90}{dyslexic} & \rotatebox{90}{L2} & \rotatebox{90}{children} & \rotatebox{90}{news} & \rotatebox{90}{test.synth} & \rotatebox{90}{dativitis}  & \rotatebox{90}{spaces} & \rotatebox{90}{commas} & \rotatebox{90}{dupl-words} & \rotatebox{90}{mood} & \rotatebox{90}{rand-noise} & \rotatebox{90}{noun-case}\\
\midrule
(not trained) & no corr.  & 83.4 & 39.1 & 47.5 & 33.3 & 76.6 & 2.9 & 74.3 & 72.3 & 65.8 & 79.0 & 65.9 & 64.6 & 31.0 \\

\midrule
(rule-based) & GC (spell.) & 84.6$^\ast$ & 44.2$^\ast$ & 47.3$^\ast$ & 38.2$^\ast$ & 89.1$^\dagger$ & 31.8 & 73.4 & 75.6 & 65.1 & 92.8 & 65.4 & 83.8 &	30.6	\\
& GC (all)  & 83.6$^\ast$ & 44.6$^\ast$ & 47.3$^\ast$ & 38.2$^\ast$ & \underline{\textbf{92.9$^\dagger$}} & 31.8 & 88.0 & 74.6 & 63.5 & 97.4 & 67.6 & 84.1 & 37.8	\\
\midrule

Synth 100k & \texttt{mB-ISEN}  & 83.4 & 44.5 & 48.4 & 39.1 & 83.3 & 89.9 & 90.0 & \textbf{98.0} & 73.8 & \underline{\textbf{98.4}} & 88.1 & 96.7 & 87.9 \\
 & \texttt{mB-ISEN+reg}  & 83.4 & 44.5 & 48.3 & 39.3 & 83.4 & 90.1 & 89.8 & 97.9 & 75.6 & \textbf{98.2} & \underline{\textbf{88.9}} & \textbf{96.8} & 88.5 \\
& \texttt{mT5} & 83.4 & 44.1  & 47.4 & 38.3 & 83.4 & 84.4 & 83.5 & 97.6 & 71.7 & 97.9 & 83.8 & 95.5 & 82.8 \\
& \texttt{ByT5} & 83.6 & 46.1 & 48.9 & 39.7 & 84.0 & \textbf{91.5} & \textbf{94.3} & \underline{\textbf{98.2}} & \textbf{76.6} & \textbf{98.2} & 83.0 & \textbf{96.8} & 88.7\\
\midrule
Error Corpora (EC) & \texttt{mB-ISEN} & 85.3 & 45.0 & 52.2 & 38.9 & 82.2 & 12.0 & 73.0 & 88.3 & 65.9 & 86.0 & 80.0 & 74.5 & 47.1  \\
& \texttt{mB-ISEN+reg} & 85.7 & 46.8 & 53.6 & 42.3 & 82.8 & 13.8 &  74.1 & 89.6 & 66.7 & 85.4 & 80.8 & 75.2 & 48.8  \\
 &  \texttt{mT5} & 84.6 & 41.7 & 46.4 & 36.8 & 78.5 & 4.4 & 73.3 & 82.2 & 64.8 & 81.9 & 65.4 & 65.5 & 30.3 \\
 &  \texttt{ByT5} & 85.9 & 46.5 & 50.6 & 43.4 & 81.8 & 20.0 & 72.2 & 90.0 & 69.1 & 89.0 & 71.2 & 75.0 & 32.7 \\

\midrule
Synth 100k + EC & \texttt{mB-ISEN} & 86.5 & 50.0 & 57.6 & 47.6 & 86.3 & 79.3 & 89.6 & 95.6 & 72.0 & 96.2 & 87.3 & 93.0 & 77.7 \\
& \texttt{mB-ISEN+reg} & 86.4 & 50.8 & 56.8 & 47.5 & 86.1 & 80.1 & 90.1 & 96.1 & 72.5 & 96.9 & 87.6 & 93.4 & 79.8 \\
& \texttt{mT5} & 86.3 & 47.4 & 52.8 &	42.9 &	85.8 & 65.0 & 75.3 & 95.3 &	69.1 &	95.6 &	80.1 & 90.4 & 57.7 \\
& \texttt{ByT5} & \textbf{87.0} & \textbf{53.4} & \textbf{58.3} & \textbf{50.1} & 87.8 & 80.5 & 93.1 & 96.6 & 72.9 & 81.3 & 86.4 & 93.5 & 81.0 \\

\midrule
\midrule
Synth 550k & \texttt{ByT5} & 84.2 & 46.3 & 49.4 & 40.9 & 84.3 &  \underline
{\textbf{95.0}} & \underline
{\textbf{96.4}} & \textbf{98.0} & \underline
{\textbf{78.9}} & 97.0 & 86.5 & \underline
{\textbf{97.2}} & \underline
{\textbf{91.4}} \\
Synth 550k + EC & \texttt{ByT5} & \underline
{\textbf{87.4}} & \underline
{\textbf{54.9}} & \underline
{\textbf{59.6}} & \underline
{\textbf{51.9}} & \textbf{89.4} & 88.5 & 93.0 & 96.8 & 75.5 & 88.6 & \textbf{88.2} & 94.8 & \textbf{90.2} \\

\bottomrule
\end{tabularx}
}

\normalfont
\caption{GLEU scores for Icelandic GEC over different model and data combinations. The two highest-scoring configurations for each subcorpus are in bold while the highest score is underlined. Scores marked with a symbol are explained in Section \ref{tab:results}.}
\label{tab:results}
\end{table*}

We consider a variety of curated and synthetic test sets to get a good overview of the differences between the byte-level and subword-level approach for GEC. For the real errors, we report scores over the \textit{IceEC.test} set, the test set from the IceEC, which contains around 5000 sentences. In contrast, the \textit{dyslexic}, \textit{L2} and \textit{children} test sets contain 500 held-out sentences each from the respective specialized error corpora described in section \ref{sec:curated} (only 100 examples were collected for the dativitis error type, a rarer occurrence in the data).
We also annotated a small dataset (163 sentences) of data from an Icelandic news outlet (\textit{news}), where each sentence contains at least one error; this is further described in Appendix \ref{sec:newstest}.

For the synthetic errors, we report GLEU scores over the \textit{test.synth} set, which contains around 4000 held-out sentences from the synthetic data. 
Furthermore, we generated test sets of synthetic examples, each containing a particular error type in each sentence (\textit{dativitis}, \textit{spaces}, \textit{commas}, \textit{dupl-words}, \textit{mood}, \textit{rand-noise}, \textit{noun-case}).
This last group of test sets was generated using source texts that, while editorial, may include other errors, just like the synthetic training examples.
The models, as they get better, learn to correct these errors as well.
This may paradoxically lower the GLEU score as the corrected output deviates from the erroneous reference.
These generated test sets still provide valuable information about what the models learn about each error type in isolation.

To understand what approach is best suited for GEC we trained the models on different data combinations and using different pre-trained models. The \texttt{Synth-100k} models are all trained for 100k updates on the same synthetic data, and the \texttt{Synth-100k-EC} models are additionally finetuned on the curated IceEC error corpus.
To provide a baseline for the GLEU scores, we also report \texttt{no\_corr} scores, where the source text is not corrected.
This gives some idea of the noise level of the test sets, with \textit{test.synth} being the noisiest and \textit{IceEC.test} containing the least noise.

The GreynirCorrect \cite{oladottir-etal-2022-developing} (GC) results were produced by applying GreynirCorrect to the test sets in its two configurations; correcting single-word errors on the one hand (\textit{spell.}) and all errors found on the other (\textit{all}).
The GC system was developed in part to focus on the error categories and error frequencies defined in the IceEC training data, and this may be reflected in the scores for the IceEC test sets (asterisked in Table \ref{tab:results}).
The \textit{news} test set ($\dagger$) was created using only sentences flagged by GC (see Appendix \ref{sec:newstest}) and is therefore heavily biased towards that system.

Models trained only on the synthetic data (\texttt{Synth 100k}) generally perform best on the synthetic error corpora and thus solve the more simple and systematic spelling errors such as mistakes in punctuation, missing white space and word duplication.
Their performance on the curated error corpora is somewhat lacking.

In contrast, models trained only on the curated error corpora (\texttt{EC}) generally produce somewhat better GLEU scores on the curated error corpora than the \texttt{Synth 100k} models, but do not generalize to the error categories presented in the synthetic test sets. They are also unable to correct multiple errors in a single sentence (\textit{test.synth}).

Training on the synthetic data and then finetuning on the curated error corpora (\texttt{Synth100k/ Synth550k+EC}) performs best on the curated errors and retains much of the performance on the synthetic test sets.
In all of these experiments, we can see that the ByT5 models generally perform better than the subword counterparts.

This is also reflected in the ERRANT scores in Appendix \ref{sec:errant}, Table \ref{tab:errant-results}, where the ByT5 models score highest overall. %

\section{Discussion \& Conclusion}
\label{sec:discussion}

Our results show that the ByT5 models are the overall high-scorers on the real-world test sets, and on most of the synthetic ones. We include finetuning results on the ByT5 model that has been trained for longer on the synthetic data (550k updates) to compare how performance improves with time. We see the GLEU scores keep going up with time, and more importantly, when taking a close look at the actual generated output, this is the model that best corrects real-world errors. This makes it the most feasible option for use in real-world scenarios. A comparison of the output of the models trained on both data sources is shown in Appendix \ref{sec:exampleoutputs}. 

An example from the test data is when the subword-tokenized model \texttt{mBART-ENIS-Synth100k+EC} incorrectly changes the name of a person from a rare name (``Láretta'') to a more common one (``Lára''). This kind of error is not seen in the byte-level model, which is quite conservative in its corrections of unknown entities. While this means ByT5 occasionally misses actual errors, we find that it is much better suited for production than a subword-level model that makes serious semantic errors. These more nuanced error correction examples may not be fully captured by the automated metrics, but are crucial for real-world use.

The subword regularization experiments are included as an alternative approach for mitigating the subword tokenization problem. The results are marginally better than the model without subword regularization when trained on the synthetic data, and the model performs better than the \texttt{ByT5-Synth100k}\ model in the case of duplicate words, which linguistically is a quite trivial task, and in more intricate mood errors. It however doesn't do any better than the \texttt{mB-ISEN-Synth100k}\ trained without subword regularization on the curated datasets, and this also holds when the model is finetuned additionally on curated data. The model finetuned on only the curated data with subword regularization (\texttt{mB-ISEN-reg-EC}) however performs consistently much better than its counterpart without subword regularization, often on par with or surpassing ByT5. This model has not seen any of the highly noised synthetic data, and thus has the most to gain from the subword noise. We speculate that this is one of the reasons we don't see more gains from adding subword regularization; the training examples are already so highly noised that there is not much to be learned from the added subword noise.

The IceEC finetuning data contain real-world errors which have been hand-corrected. These texts are somewhat different from the highly noised training examples with synthetic errors, have fewer errors on average and are more varied as they are naturally occurring. They also include stylistic edits from the reviewers, which improve the text's fluency, but in those cases the original is not necessarily incorrect as per the language standard. With these differences in mind, we expect the models to have to forget some of the synthetic error correction task in order to adapt to this ``new'' denoising task. We see this happen in the mBART-ENIS finetuning on the curated data, and to a lesser extent in the ByT5 finetuning. The denoising task performance on the synthetic errors from the previous step has in part been lost, which is expected, since some of these errors are not particularly common in real texts. 

For the more grammatically complex error-types in the synthetic data (dativitis and changes to noun cases and verb moods), we find that the mBART-ENIS trained on synthetic data generally does well; for some subsets even surpassing the ByT5 counterpart that was finetuned on curated corpora. We suspect that this has to do with the linguistic knowledge the model has already gained during its pretraining on Icelandic texts, as explained in Appendix \ref{sec:mbart}. The ByT5 model that was trained for longer however manages to surpass it on the \emph{mood} error type, indicating that it is still adapting to the Icelandic language, alongside its primary denoising task.

The models trained on only the finetuning data perform the worst throughout. The results show that they do not manage to correct the synthetic categories much beyond the baseline, except for mBART-ENIS in some cases. We expect that this has to do with their extra knowledge of Icelandic and the denoising objective used in the synthetic error correction finetuning. The results for these models on the curated in-domain test sets are in fact mostly on par with the models finetuned on the synthetic data only. Looking at the generated output, we see that the error types these models correct are not the same as those that the synthetic-only models are able to correct, which is expected, as they are trained on different data.

We conclude that adopting a byte-level approach rather than a subword approach leads to best results for the task of GEC, at the very least in the case of a morphologically rich language such as Icelandic. Finally, we find that the optimal way of capturing a wide range of errors is to train on a combination of synthetic and curated data, particularly when the curated data is limited.

\section*{Limitations}
Potential limitations to our work can mainly be attributed to two factors: 1) the fact that we run our experiments using the Icelandic language, and 2) inherent biases in the corpora we use.

Icelandic is North Germanic language (along with Faroese, Norwegian, Danish and Swedish). As such, it is both Germanic and Indo-European. While we are fairly confident that our results hold for these languages, different results may hold for other languages, particularly those not using Latin script or those using logograms, such as Chinese characters.

The curated datasets we use only represent a fairly small proportion of all possible demographics and users of the Icelandic language. In particular, annotations are performed by a handful of university students, bringing in their biases to the annotated data. Even so, the data should serve well to compare the relative differences.

The resources we use to develop the models consist of a few high-performing GPUs. While these are powerful, this is a relatively low requirement compared to many industry or academic use cases.

Finally, it is worth re-iterating that the ByT5 model we use is slow compared to subword-based models for texts of similar length. Inference in our setting was around 2.3x slower on average than for mT5. As such, production use of these methods may be better suited to offline processing, particularly for longer documents.

\section*{Ethics Statement}
While we do not believe the data we use to train the error-correcting models to be sensitive, the models can be applied in sensitive settings where an incorrect edit may cause an issue. As such, corrections may introduce both stylistic or semantic changes based on either the biases found in the pretrained models or the curated error corpora.

In particular, we have noticed a bias in the subword-based models for entities, such as locations, being overcorrected to a different entity if there is a spelling mistake in the input.

The stylistic changes found in the curated Icelandic corpora may reflect on the socio-economic background of the annotators and writers of the data. While we don't believe this to be a large issue in this particular setting, one can easily imagine this to be more complex in regions where language use is connected to disputes or oppression. As such, a text correction or improvement tool could be used to homogenize discourse or otherwise limit freedom of expression, knowingly or unknowingly.

\section*{Acknowledgements}
We thank the Icelandic Language Technology Program \cite{nikulasdottir-etal-2020-language}. It has enabled the authors to focus on work in Icelandic NLP. Snæbjarnarson was partially funded by the Pioneer Centre for AI, DNRF grant number P1, during the time of this work. Finally, we thank the anonymous reviewers for their helpful feedback.

\bibliography{anthology,custom_fix}
\bibliographystyle{acl_natbib}

\appendix

\section{mBART-ENIS training}
\label{sec:mbart}
We continued the pretraining of the multilingual BART25 model using texts from various sources in Icelandic and English. The Icelandic text used was the Icelandic Common Crawl corpus (IC3) \cite{snaebjarnarson-etal-2022-warm}, IGC \cite{steingrimsson-2018-very}, papers published in the Icelandic Medical Journal (\url{https://hirsla.lsh.is}), text extracted from Icelandic student theses (\url{https://skemman.is}), and Icelandic e-books in the public domain (\url{https://rafbokavefur.is}). English data used was the Newscrawl 2019-2020 dataset \cite{tiedemann-2012-parallel}, English Wikipedia and the Books3 corpus \cite{soskkobayashi2018bookcorpus}.

The goal of the pretraining is that the model already has some knowledge of the Icelandic language. English is included as we speculate that it is beneficial to continue including a language used in the earlier training, making the model a better starting point for other use cases such as machine translation and cross-lingual transfer tasks.

Due to the English data outnumbering the Icelandic data, we upsampled the Icelandic data by about a factor of 6.5, resulting in a sample language probability of 55\% English and 45\% Icelandic. 

The model has 354M non-embedding parameters, and 256M parameters for embeddings. Note that not all of them are used when only training on Icelandic and English. The model was trained for 316k updates with an effective batch size of 44k tokens per update, 5k warmup steps, a learning rate of \textrm{7.5e-5} and a dropout of 0.2. Other hyperparameters, such as for noising, were the same as for the original mBART model. The training took approximately 18 A100 days but was not continued until convergence due to computational constraints.

\section{News test set collection and annotation}
\label{sec:newstest}
We collected a small test set of erroneous sentences by running news articles from the RÚV (National Icelandic Broadcasting Service) website through the open-source GreynirCorrect \cite{oladottir-etal-2022-developing} spelling and grammar correction tool for Icelandic. We filtered out the sentences flagged as containing errors, and manually chose 163 sentences that contained only a single obvious error, for evaluation clarity. The sentences were then hand-corrected by a linguist and used as an additional test set in our experiments. (Since GreynirCorrect was used to find the erroneous sentences, this test set is unreliable for evaluating that particular system's performance.)

\section{Icelandic-specific grammatical noise}
\label{sec:appendixis}
We generated various grammatical errors to create our synthetic error corpus. Icelandic has four grammatical cases; we swapped those randomly in nouns, producing ungrammatical sentences of a type commonly seen in learner texts. We also changed the mood of verbs from the subjunctive to the indicative, a variation often seen for both native speakers and learners. Another common variation, which is more or less accepted in informal language but still discouraged in formal language and written texts, is the so-called ``dativitis'', i.e., the use of the dative case instead of the accusative or nominative with certain verbs with oblique subjects, such as ``mér (dat) hlakkar'' (``I look forward to'') instead of ``ég (nom) hlakka'', or ``Páli (dat) langar'' (``Páll wants'') instead of ``Pál (acc) langar''. This modification was produced using the Greynir engine by extracting and modifying whole nominal clauses.

We used available resources to apply realistic misspellings to single words -- these are lists of common misspellings and their corrections. We located the corresponding correctly spelled words in the corpus and substituted their misspelled variants from the error lists. The error lists are sourced from nonwords and misspellings in IceSQuER \cite{arnardottir-2020-icelandic}, the IceEC \cite{arnardottir-2021-creating}, The Database of Modern Icelandic Inflection \cite{bjarnadottir-2019-database}, and GreynirCorrect \cite{oladottir-etal-2022-developing}.

\section{Example outputs}
\label{sec:exampleoutputs}
We selected three sentences at random from the dyslexia subcorpus of the IceEC. These sentences, their corrected reference and the respective corrected outputs from various model variants can be seen in Table \ref{tab:errorexamples}. The examples show some of the paraphrasing and stylistic changes that appear in the IceEC, which can be substantial, and that we don't expect the neural models to mimic.

\begin{table*}
\begin{tabularx}{\textwidth}{l X}
\toprule
\textbf{Model} & \textbf{Text}\\
\midrule

Original & Ef notandi valdi að \textit{\textcolor{red}{ítta}} á rétt kemur upp þessi síða þar sem notandi getur \textcolor{red}{\textit{seð} \textit{nánar}} um \textcolor{red}{\textit{réttin} \textit{með innihaldsefnun}} og \textcolor{red}{\textit{næringaupplýsingum}}.
 \\
Reference & Ef notandi valdi að \textcolor{teal}{\underline{ýta}} á rétt kemur upp þessi síða þar sem notandi getur \textcolor{teal}{\underline{séð nánari upplýsingar}} um \textcolor{teal}{\underline{réttinn, innihaldsefni og næringarupplýsingar}}.
\\
\texttt{mB-ISEN-Synth100k+EC} & Ef notandi valdi að \textcolor{teal}{\underline{ýta}} á rétt kemur upp þessi síða þar sem notandi getur \textcolor{teal}{\underline{séð}} \textcolor{red}{\textit{nánar}} um \textcolor{teal}{\underline{réttinn}} með \textcolor{red}{\textit{innihaldsefnunum}} og \textcolor{teal}{\underline{næringarupplýsingum}}.
 \\
\texttt{mT5-Synth100k+EC} & Ef notandi valdi að \textcolor{teal}{ýta} á rétt kemur upp þessi síða þar sem notandi getur \textit{\textcolor{red}{seð nánar}} um \textcolor{teal}{\underline{réttinn}} með \textit{\textcolor{red}{innihaldsefnun}} og \textit{\textcolor{red}{næringaupplýsingum}}.
 \\
\texttt{ByT5-Synth100k+EC} & Ef notandi valdi að \textcolor{teal}{\underline{ýta}} á rétt kemur upp þessi síða þar sem notandi getur \textcolor{teal}{\underline{séð}} \textit{\textcolor{red}{nánar}} um \textcolor{teal}{\underline{réttinn}} með \textcolor{teal}{\underline{innihaldsefnum}} og \textcolor{teal}{\underline{næringarupplýsingum}}.
 \\
\texttt{ByT5-Synth550k+EC} & Ef notandi velur að \textcolor{teal}{ýta} á rétt kemur upp þessi síða þar sem notandi getur \textcolor{teal}{\underline{séð}} \textit{\textcolor{red}{nánar}} um \textcolor{teal}{\underline{réttinn}} með \textit{\textcolor{red}{innihaldsefni}} og \textcolor{teal}{\underline{næringarupplýsingum}}.
 \\
\midrule
Original & En aftur á móti hefur sá hópur sem reykir \textit{\textcolor{red}{marjúana}} daglega \textit{\textcolor{red}{eining}} stækkað, en dregið úr áfengisneyslu.
 \\
Reference & Sömuleiðis hefur dregið úr áfengisneyslu. Aftur á móti hefur sá hópur sem reykir \textcolor{teal}{\underline{marijúana}} daglega \textcolor{teal}{\underline{einnig}} stækkað.
\\
\texttt{mB-ISEN-Synth100k+EC} & En aftur á móti hefur sá hópur sem reykir \textit{\textcolor{red}{marjúana}} daglega \textit{\textcolor{red}{eining}} stækkað, en dregið úr áfengisneyslu.
 \\
\texttt{mT5-Synth100k+EC} & En aftur á móti hefur sá hópur sem reykir \textit{\textcolor{red}{marjúana}} daglega \textit{\textcolor{red}{eining}} stækkað, en dregið úr áfengisneyslu.
 \\
\texttt{ByT5-Synth100k+EC} & Aftur á móti hefur sá hópur sem reykir \textit{\textcolor{red}{marjúana}} daglega \textcolor{teal}{\underline{einnig}} stækkað, en dregið úr áfengisneyslu.
 \\
\texttt{ByT5-Synth550k+EC} & Aftur á móti hefur sá hópur sem reykir \textcolor{teal}{\underline{marijúana}} daglega \textcolor{teal}{\underline{einnig}} stækkað, en dregið úr áfengisneyslu.
 \\
\midrule
Original &  Það \textit{\textcolor{red}{koma}} alltaf sama villan sem má sjá \textit{\textcolor{red}{\_}} mynd 1.
 \\
Reference & Það \textcolor{teal}{\underline{kom}} alltaf sama villan sem má sjá \textcolor{teal}{\underline{á}} mynd 1.
\\
\texttt{mB-ISEN-Synth100k+EC} & Það \textit{\textcolor{red}{kemur}} alltaf sama villan sem má sjá \textcolor{teal}{\underline{á}} mynd 1.
 \\
\texttt{mT5-Synth100k+EC} & Það \textit{\textcolor{red}{koma}} alltaf sama villan sem má sjá \textcolor{teal}{\underline{á}} mynd 1.
 \\
\texttt{ByT5-Synth100k+EC} & Það \textit{\textcolor{red}{koma}} alltaf sama villan sem má sjá \textcolor{teal}{\underline{á}} mynd 1.
 \\
\texttt{ByT5-Synth550k+EC} & Það \textcolor{teal}{\underline{kom}} alltaf sama villan sem má sjá \textcolor{teal}{\underline{á}} mynd 1.
 \\

\bottomrule
\end{tabularx}
\caption{Example outputs from selected models, on evaluation data sourced from the IceEC dyslexia corpus. Red indicates an error, and teal a correct word or phrase in the context of the sentence. Model outputs may not correspond to the hand-corrected reference sentence, but still be a quality correction of the original.}
\label{tab:errorexamples}
\end{table*}

\section{ERRANT results}
\label{sec:errant}

Table \ref{tab:errant-results} presents ERRANT scores for the test sets. The data is tokenized and converted to M2 format, then evaluated using the original ERRANT scorer developed by \citet{bryant-etal-2017-automatic}.

\begin{table*}
\small
\resizebox{\textwidth}{!}{%
\begin{tabularx}{1.15\textwidth}{l|l||lllll|llllllll}
\toprule
 \multicolumn{1}{c}{} & & \multicolumn{5}{c|}{\sc{Curated Error Corpora}} & \multicolumn{8}{c}{\sc{Synthetic Error Corpora}} \\ 
\midrule 
\sc{Training Data} & \sc{Model} & \rotatebox{90}{IceEC.test} &  \rotatebox{90}{dyslexic} & \rotatebox{90}{L2} & \rotatebox{90}{children} & \rotatebox{90}{news} & \rotatebox{90}{test.synth} & \rotatebox{90}{dativitis}  & \rotatebox{90}{spaces} & \rotatebox{90}{commas} & \rotatebox{90}{dupl-words} & \rotatebox{90}{mood} & \rotatebox{90}{rand-noise} & \rotatebox{90}{noun-case}\\
\midrule
(not trained) & GC (spell.) & $0.36^\ast$ & $0.35^\ast$ & $0.20^\ast$ & $0.37^\ast$ & $\textbf{0.78}^\dagger$ & 0.75 & 0.05 & 0.54 & 0.00 & 0.86 & 0.06 & 0.79 & 0.04\\
(rule-based) & GC (all) & $0.32^\ast$ & $0.35^\ast$ & $0.22^\ast$ & $0.40^\ast$ & $0.75^\dagger$ & 0.75 & 0.72 & 0.47 & 0.00 & 0.87 & 0.34 & 0.74 & 0.40\\
\midrule

Synth 100k
& \texttt{mB-ISEN} & 0.48 & 0.51 & 0.50 & 0.54 & 0.69 & 0.92 & 0.66 & 0.79 & 0.40 & 0.85 & 0.77 & 0.84 & 0.87\\
& \texttt{mB-ISEN+reg} & 0.46 & 0.52 & 0.48 & 0.55 & 0.67 & 0.92 & 0.70 & 0.82 & 0.40 & 0.87 & \textbf{0.78} & 0.85 & 0.88\\
& \texttt{mT5} & 0.26 & 0.35 & 0.20 & 0.35 & 0.57 & 0.95 & 0.66 & 0.89 & 0.39 & 0.90 & 0.77 & 0.91 & 0.92\\
& \texttt{ByT5} & 0.29 & 0.40 & 0.23 & 0.37 & 0.58 & 0.97 & 0.85 & \textbf{0.91} & 0.59 & \textbf{0.93} & 0.76 & \textbf{0.92} & \textbf{0.93}\\
\midrule
Error Corpora (EC)
& \texttt{mB-ISEN} & 0.40 & 0.37 & 0.35 & 0.36 & 0.53 & 0.51 & 0.00 & 0.75 & 0.13 & 0.70 & 0.71 & 0.61 & 0.63\\
& \texttt{mB-ISEN+reg} & 0.44 & 0.41 & 0.41 & 0.47 & 0.57 & 0.54 & 0.07 & 0.74 & 0.20 & 0.65 & 0.71 & 0.61 & 0.65\\
& \texttt{mT5} & 0.31 & 0.17 & 0.13 & 0.24 & 0.29 & 0.26 & 0.00 & 0.66 & 0.00 & 0.50 & 0.03 & 0.21 & 0.00\\
& \texttt{ByT5} & 0.45 & 0.38 & 0.30 & 0.45 & 0.50 & 0.65 & 0.02 & 0.77 & 0.27 & 0.76 & 0.44 & 0.60 & 0.15\\

\midrule
Synth 100k + EC
& \texttt{mB-ISEN} & 0.48 & 0.51 & 0.50 & 0.54 & 0.69 & 0.92 & 0.66 & 0.79 & 0.40 & 0.85 & 0.77 & 0.84 & 0.87\\
& \texttt{mB-ISEN+reg} & 0.46 & 0.52 & 0.48 & 0.55 & 0.67 & 0.92 & 0.70 & 0.82 & 0.40 & 0.87 & \textbf{0.78} & 0.85 & 0.88\\
& \texttt{mT5} & 0.47 & 0.44 & 0.38 & 0.47 & 0.69 & 0.89 & 0.30 & 0.83 & 0.29 & 0.90 & 0.70 & 0.84 & 0.76\\
& \texttt{ByT5} & 0.52 & 0.56 & 0.52 & 0.57 & 0.70 & 0.92 & 0.78 & 0.83 & 0.45 & 0.45 & \textbf{0.78} & 0.86 & 0.88\\
\midrule
\midrule
Synth 550k & \texttt{ByT5} & 0.31 & 0.42 & 0.25 & 0.39 & 0.61 & \textbf{0.98} & \textbf{0.87} & 0.90 & \textbf{0.63} & 0.88 & 0.75 & \textbf{0.92} & 0.91\\
Synth 550k + EC & \texttt{ByT5} & \textbf{0.54} & \textbf{0.57} & \textbf{0.54} & \textbf{0.58} & 0.74 & 0.95 & 0.75 & 0.82 & 0.51 & 0.72 & \textbf{0.78} & 0.85 & 0.90\\
\bottomrule
\end{tabularx}
}

\normalfont
\caption{ERRANT span-based $F\textsubscript{0.5}$ scores calculated over different model and data combinations. The best scoring setup per error subcorpus is in bold. Scores marked with a symbol are explained in Section \ref{tab:results}.}
\label{tab:errant-results}
\end{table*}

\end{document}